\pdfoutput=1

\documentclass[11pt]{article}

\usepackage{acl}

\usepackage{times}
\usepackage{latexsym}
\usepackage{enumitem}
\usepackage[T1]{fontenc}

\usepackage[utf8]{inputenc}

\usepackage{microtype}

\usepackage{inconsolata}

\usepackage{graphicx}
\usepackage{multirow}
\usepackage{bm}
\usepackage{algorithm}
\usepackage{algpseudocode}
\usepackage{amsmath} 
\usepackage{mathtools}
\usepackage{hyperref}
\usepackage{booktabs}
\usepackage{amssymb}
\usepackage{microtype}
\title{M\textsuperscript{3}HG: Multimodal, Multi-scale, and Multi-type Node Heterogeneous Graph for Emotion Cause Triplet Extraction in Conversations}

\author{
    Qiao~Liang\quad Ying~Shen\thanks{$\ $  Corresponding authors.}\quad Tiantian~Chen\quad Lin~Zhang \\
    Tongji University, Shanghai, China\\
    \texttt{\{2333091, yingshen, 2111287, cslinzhang\}@tongji.edu.cn}\thanks{\noindent 
This work was supported in part by the National Natural Science Foundation of China under Grant 62476202 and 62272343, in part by the Fundamental Research Funds for the Central Universities.}
}

\begin{document}

\maketitle

\begin{abstract}
Emotion Cause Triplet Extraction in Multimodal Conversations (MECTEC) has recently gained significant attention in social media analysis, aiming to extract emotion utterances, cause utterances, and emotion categories simultaneously.
However, the scarcity of related datasets, with only one published dataset featuring highly uniform dialogue scenarios, hinders model development in this field.
To address this, we introduce \textbf{MECAD}, the first multimodal, multi-scenario MECTEC dataset, comprising 989 conversations from 56 TV series spanning a wide range of dialogue contexts. 
In addition, existing MECTEC methods fail to explicitly model emotional and causal contexts and neglect the fusion of semantic information at different levels, leading to performance degradation.
In this paper, we propose M\textsuperscript{3}HG, a novel model that explicitly captures emotional and causal contexts and effectively fuses contextual information at both inter- and intra-utterance levels via a multimodal heterogeneous graph.
Extensive experiments demonstrate the effectiveness of M\textsuperscript{3}HG compared with existing state-of-the-art methods.
The codes and dataset are available at  \url{https://github.com/redifinition/M3HG}.
\end{abstract}

\section{Introduction}

Emotion Cause Analysis in Conversations (ECAC) aims at identifying emotions and their causes in conversations, which is a crucial research field in natural language processing~\cite{li2022ecpec, wang2023generative}.
However, most of ECAC research~\cite{li2022ecpec, wang2023generative, ECQED, MPEG} only focuses on the textual contexts, overlooking other modalities~\cite{soleymani2017survey}.

\begin{figure*}
\centering
  \includegraphics[width=0.9\textwidth]{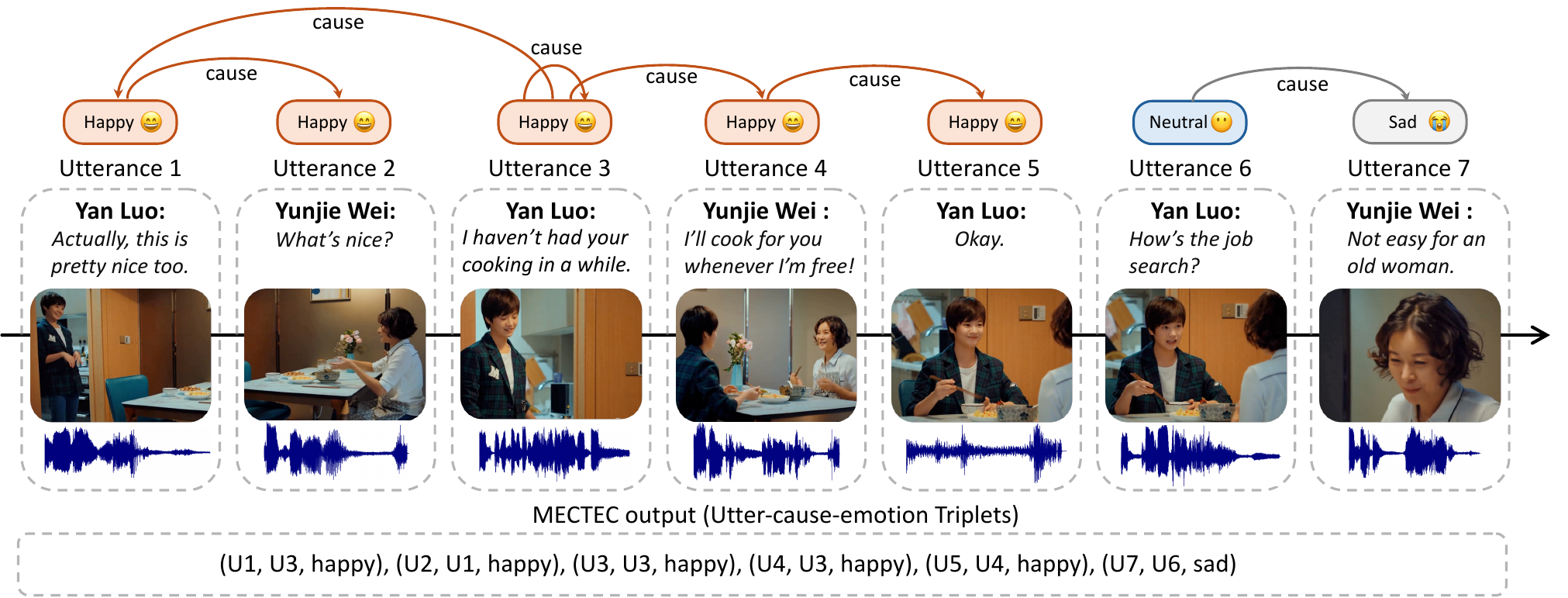}
  \caption{An example of the MECTEC task. Each utterance contains three different modalities - text, audio, and video. Arrows represent causal relationships that link the cause utterances to the corresponding emotion utterances. The dashed box at the bottom lists all the <utter-cause-emotion> triplets identified in this example.}
  \label{fig:introduction figure}
  \vspace{-2mm}
\end{figure*}

To address this limitation, \citet{ecf} proposed a new task called Multimodal Emotion Cause Triplet Extraction in Conversations (MECTEC).
The task aims to simultaneously identify the emotion utterance, the corresponding cause utterances, and the emotion category (i.e., the \textit{utter-cause-emotion} triplet) from a conversation containing three modalities: text, audio, and video.
Figure~\ref{fig:introduction figure} illustrates a multimodal conversation between a mother and daughter.
In this example, there are six non-neutral utterances, and consequently, six utter-cause-emotion triplets are identified.
MECTEC differs from ECAC in 1) multimodal contexts (i.e., text, audio, and video) resulting in more complex emotional expression, and 2) multi-scale semantic information from overall conversation and utterance features like intonation and facial expressions, which pose significant challenges.

Another major challenge in MECTEC is the scarcity of datasets.
While numerous text-based datasets exist for ECAC, only one dataset, namely the ECF dataset~\cite{ecf}, is specifically designed for MECTEC.
However, the videos in ECF are all from the \textit{Friends} TV series with restricted speakers and scenarios, hindering MECTEC model development.
Therefore, in this work, a new multimodal, multi-scenario MECTEC dataset, namely \textbf{MECAD}, is constructed.
To the best of our knowledge, it is the first of its kind and will greatly facilitate research in this field.

Constrained by the limited dataset,  existing MECTEC models have various deficiencies.
~\citet{ecf} proposed a two-stage architecture that predicts emotion and cause utterances separately.
However, this approach is computationally intensive and prone to error accumulation.
Therefore, recent studies ~\cite{UniMEEC, wang2023generative, li2024multimodal} propose one-stage architectures using graph neural networks or prompt engineering to extract utter-cause-emotion triplets.
However, these methods do not explicitly extract specific contexts related to emotions and their causes.
According to emotion attribution theory~\cite{weiner1985attributional}, the relationships of emotions and their causes are revealed by specific contexts, such as emotional words in texts, and intonations in audio and video conversations.
For example, in Utterance 3 in Figure~\ref{fig:introduction figure},  a pleasant facial expression indicates happiness, while “haven't had your cooking” and a happy tone reveal the cause.
The example illustrates that emotions and their causes depend on contextual cues across multiple modalities, highlighting the necessity of explicitly modeling their specific contexts.

In addition, previous work~\cite{ecf, UniMEEC, wang2023generative, li2024multimodal, wei2020effective} fail to effectively identify the \textbf{cause utterances occurring after emotion utterances}. 
For example, in Utterance 1 in Figure 1, the reason why Luo is happy cannot be obtained only from the historical context of Utterance 1.
To find out the real cause of emotion in Utterance 1, the whole conversation should be scrutinized, which is overlooked by previous work.

Furthermore, existing models~\cite{ecf, UniMEEC, wang2023generative, li2024multimodal} fail to adequately extract semantic information at different scales.
As shown in Figure~\ref{fig:introduction figure}, the semantic information that reveals the relationship of an utterance and its cause not only resides in inter-connections between utterances but also resides in the intra-content of each utterance.
Therefore, it's essential to comprehensively integrate semantic information in different scales during modality fusion.

To solve the aforementioned problems, we propose an MECTEC model based on the multimodal, multi-scale, and multi-type node heterogeneous graph, named \textbf{M\textsuperscript{3}HG}.
M\textsuperscript{3}HG accurately extracts emotion and cause-related contexts and fuses multimodal, multi-scale semantic information using multimodal heterogeneous graph attention network (HGAT) with multi-type nodes.

Our contributions can be summarized as follows:
\begin{itemize}[itemsep=0pt,parsep=0pt,topsep=0pt,partopsep=0pt]
  \item The first Chinese multi-scenario MECTEC dataset, \textbf{MECAD}, and an online sentiment data annotation toolkit are constructed.
  The dataset consists of 989 conversations with 10,519 utterances annotated with important information such as emotion labels, their causes, and types of emotional causes.
  It will greatly benefit the development of models in the MECTEC and related fields.
  \item An efficient MECTEC model, namely M\textsuperscript{3}HG, is proposed to identify utter-cause-emotion triplets from multimodal conversations.
  It explicitly extracts specific emotion and cause-related contexts to find connections between emotions and causes.
  Besides, it fully integrates semantic information from inter and intra-utterance levels to enhance the model's predictive ability. 
  \item Extensive experiments are performed to verify the performance of our proposed model and other state-of-the-art models on MECAD and ECF datasets.
  Experimental results reveal that M\textsuperscript{3}HG outperforms its counterparts, which demonstrates the effectiveness of our model.
\end{itemize}

\section{Related Works}
\textbf{Emotion Cause Analysis in Conversations.}
Most existing studies on ECAC focus on Causal Emotion Entailment (CEE) and Emotion Cause Pair Extraction in Conversations (ECPEC).
CEE aims to identify which cause utterances trigger the non-neutral emotions of the target utterances.
Since CEE assumes emotion utterances are given, most related work~\cite{RECCON, li2022neutral, zhang2022tsam, gu2023page} viewed CEE as an utterance classification problem.
However, because emotions of utterances are often unknown in real-world conversations, \citet{li2022ecpec} proposed the ECPEC task which additionally predicts emotions for the target utterances.
Subsequent work~\cite{wang2023generative, zhao2023knowledge} has incorporated commonsense knowledge into GATs to improve the model's semantic understanding of emotions and causes, achieving better performance.
Besides, some methods~\cite{ECPE-2D, ECPE-MLL, wei2020effective, UECA-Prompt} from models in the Emotion cause Pair Extraction (ECPE) field are also adapted for the ECPEC task.

\noindent \textbf{Multimodal Emotion Cause Triplet Extraction in Conversations.} In recent years, multimodal conversation scenarios on social media platforms have grown significantly, as more individuals share their lives and express emotions through live streaming and various online chats.
To advance emotion cause analysis in multimodal conversation scenarios, \citet{ecf} introduced the MECTEC task and released the ECF dataset.
However, few solutions have been proposed for this recently introduced task.
\citet{li2024multimodal} incorporated emotion transition information into emotion-cause pair extraction using a novel labeling constraint, while \citet{UniMEEC} fused semantic information across modalities via prompt engineering.
These methods treat multimodal fusion and contextual information extraction for emotional causes as separate processes.
Furthermore, they fail to effectively integrate semantic information across different scales, which significantly hampers the overall performance of models in the MECTEC task.
To address these issues, we propose a model that fully integrates multi-scale semantic information from different modalities, preventing the loss of contextual information during fusion and improving triplet extraction accuracy.

\noindent\textbf{Datasets for the ECAC Task.}
\begin{table}
  \small
  \caption{A summary of datasets for ECAC task. T, A, V stand for text, audio and video respectively.}
  \label{tab:related-dataset}
  \centering
  \begin{tabular}{l|c|c|c}
    \toprule
    Dataset & Modalities & Sources &  \# Instances\\
    \midrule
    RECCON & T & Act and Daily  & 11,769 \\
    ConvECPE & T & Act  & 7,433 \\
    \midrule
    ECF & T,A,V & TV \textit{Friends}  & 13,509 \\
    \textbf{MECAD} & T,A,V & 56 TV series &  10,516 \\
  \bottomrule
  \end{tabular}
  \vspace{-4mm}
\end{table}
Table \ref{tab:related-dataset} summarizes popular datasets in ECAC.
\citet{RECCON} introduced the RECCON dataset for the ECAC task, and \citet{li2022ecpec} extended it by building the ConvECPE dataset.
Given the multimodal nature of conversations, \citet{ecf} developed the ECF dataset for MECTEC.
However, all scenes in ECF are drawn from the \textit{Friends}, limiting the diversity of conversation scenarios and contents.

\section{Proposed MECAD Dataset}
To facilitate the research in MECTEC and other related fields, we constructed a multi-scenario MECTEC dataset called MECAD.
Compared with ECF~\cite{ecf}, MECAD has more diverse conversation scenarios.
In addition to labeling emotion categories and their causes for each utterance, we also categorized the types of emotion causes (e.g., \textit{event}, \textit{expression}) and the modality of annotation (i.e., \textit{text}, \textit{audio}, or \textit{video}) to support future studies in multimodal emotion cause analysis.

We selected the publicly available M\textsuperscript{3}ED~\cite{M3ED} dataset as our data source, which contains 990 segments from 56 Chinese TV series.
However, M\textsuperscript{3}ED dataset only contains conversation scripts, audios, and screenshots, lacking corresponding videos.
Therefore, we endeavored to collect the corresponding video segments based on the conversation timestamps provided by M\textsuperscript{3}ED.
We concatenated sentences to form 989 multimodal conversations with 10,516 full utterances.

We invited 10 Chinese graduate students majored in Psychology to annotate the corresponding cause utterances, the types of emotion causes and the modal cues of annotations in the conversations. 
To obtain high-quality annotations, we designed detailed guidelines based on previous studies~\cite{dirven1997emotions, steptoe2009emotional}, trained the volunteers, and tested them with annotation cases.
Only those passing the test participated in the final annotation process.
Each volunteer was paid \$50 for their annotations.
Then, we randomly assigned three qualified annotators for each conversation.
If divergence exists among annotations from different volunteers, the final annotation for the utterance is determined by majority voting.
Two strategies were used to review and revise incorrect annotations:
1) Annotation consistency among the three annotators for each TV series is calculated.
For series with low consistency, the annotators rechecked and revised their labels as needed.
2) If disagreements remained, a fourth annotator was invited to relabel the utterances and make the final decision.

To enhance annotation efficiency and accuracy, we developed an online multimodal conversation emotion cause annotation tool.
The interface of the annotation tool is shown in Figure~\ref{fig:annotation toolkit} in Appendix~\ref{Appendix-toolkit}.
This tool is highly reusable and user-friendly, making it ideal for related research in the future.

We use Cohen’s Kappa~\cite{cohens-kappa} to assess pairwise agreement and Fleiss’s Kappa~\cite{fless-kappa} for overall consistency among annotators.
The Cohen’s Kappa results are in Appendix~\ref{appendix-b}, and the Fleiss’s Kappa score of 0.6932 exceeds the threshold of 0.61~\cite{landis1977measurement}, confirming the statistical reliability of our annotations.

The dataset statistics and detailed analysis of MECAD are presented in Figure~\ref{fig:annotation result} in Appendix~\ref{appendix-a}.
MECAD provides solid support for assessing the performance and generalization capabilities of MECTEC models in broader scenarios.
\section{\texorpdfstring{Framework of Proposed M\textsuperscript{3}HG}{Framework of Proposed M3HG}}

\begin{figure*}
\centering
  \includegraphics[width=1\textwidth]{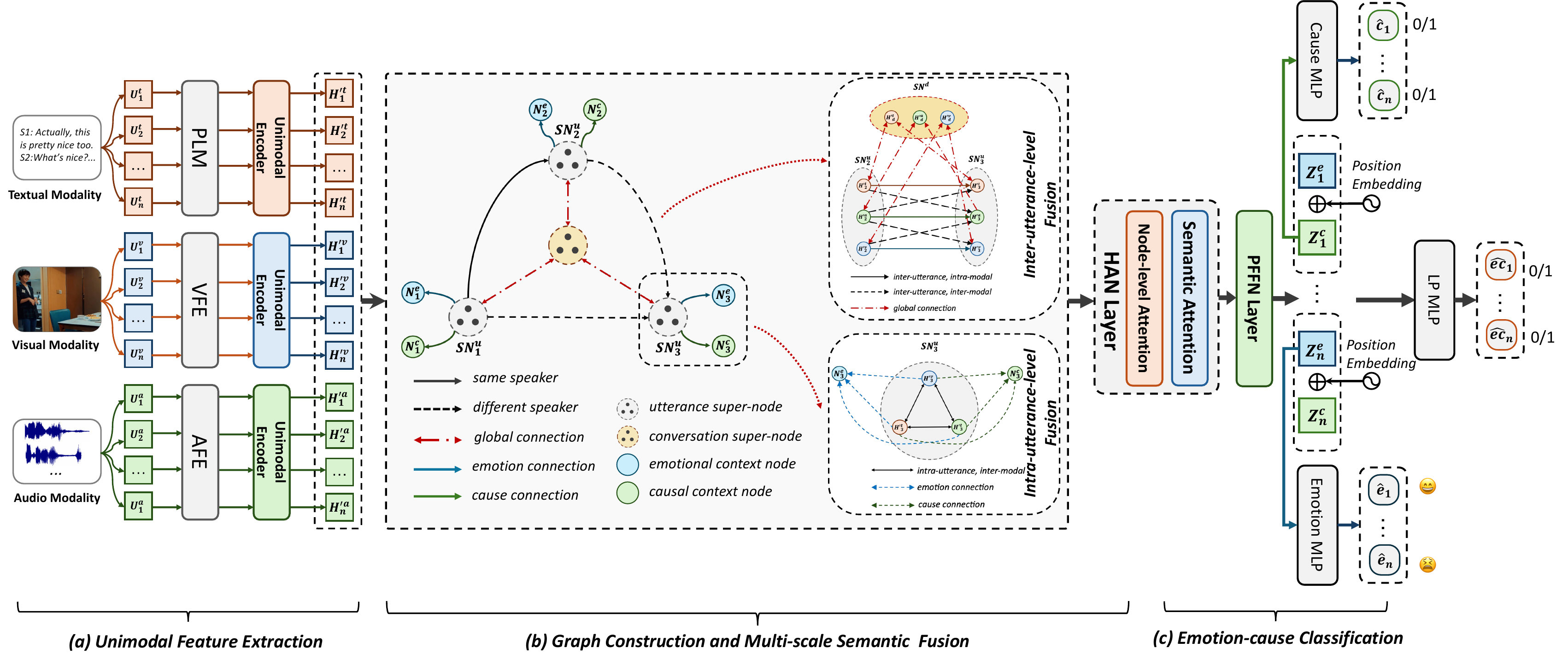}
  \caption{The framework of proposed M\textsuperscript{3}HG. It consists of three main components: unimodal feature extraction, graph construction and multi-scale semantic fusion, and emotion-cause classification.}
  \label{fig:overallframework}
\end{figure*}

\subsection{Task Definition}
Given a conversation $C = \{(S_i, \bm{U}_i)\}_{1\leq i\leq n}$, where $S_i$ denotes the speaker of the $i$-th utterance $\bm{U}_i$, $n$ denotes the length of the conversation $C$, $\bm{U}_i = \{U_i^t, U_i^a, U_i^v\}$, and $t$, $a$, $v$ are the text, audio and video modality, respectively.
The goal of MECTEC is to identify all the utter-cause-emotion triplets from the conversation $C$:
\begin{equation}
\mathcal{P} = \{(\bm{U}_j^e, \bm{U}_j^c, y_j^e)\},
\end{equation}
where $\bm{U}_j^e$ is the $j$-th utterance with emotion $y_j^e$, $\bm{U}_j^c$ is the corresponding cause utterance, and $y_j^e\in \{\textit{Anger}, \textit{Disgust}, \textit{Fear}, \textit{Joy}, \textit{Sadness}, \textit{Surprise}\}$~\cite{ekman1992argument}.

\subsection{Model Overview}
M\textsuperscript{3}HG is an end-to-end (E2E) MECTEC model, as illustrated in Figure~\ref{fig:overallframework}.
It consists of four key components: \textit{unimodal feature extraction}, \textit{graph construction}, \textit{multi-scale semantic fusion}, and \textit{emotion-cause classification}.

In \textit{unimodal feature extraction},  M\textsuperscript{3}HG extracts local contextual representations for each utterance using modality-specific feature extractors and unimodal encoders.
In \textit{graph construction}, M\textsuperscript{3}HG constructs a conversation interaction graph using these feature representations to explicitly model the emotion and cause-related contexts.
In \textit{multi-scale semantic fusion}, M\textsuperscript{3}HG combines semantic information at different scales within the conversation interaction graph to produce a comprehensive feature representation of both emotion and cause contexts.
In \textit{emotion-cause classification}, emotion and cause contextual representations are concatenated and used to extract utter-cause-emotion triplets with with position embedding.

\subsection{Unimodal Feature Extraction}
First, we utilize SA-RoBERTa~\cite{SA-RoBERTa}, Wav2Vec2~\cite{wav2vec2}, and DenseNet~\cite{DenseNet} to extract three feature representations $\bm{E}^t$, $\bm{E}^a$, and  $\bm{E}^v$, from text, audio, and video, respectively, where $\bm{E}^t \in \mathbb{R}^{n \times d_t}$, $\bm{E}^a \in \mathbb{R}^{n \times d_a}$, and  $\bm{E}^v \in \mathbb{R}^{n \times d_v}$, and  $d_t$, $d_a$, $d_v$ represent dimensions of the hidden layer representations of the three modalities.
The extraction process is described in Appendix~\ref{appendix-c.1}.

Then, we encode each feature representation within an unimodal local context.
For text, we apply multi-head self-attention~\cite{self-attention} to $\bm{E}^t$ to capture local contextual information, resulting in $\bm{H}^t$.
For $\bm{E}^a$ and $\bm{E}^v$, we use a GRU-based network~\cite{CFN-ESA} to extract local context by leveraging the RNN structure’s capability to handle temporal features, which is expressed as:
\begin{equation}
\begin{aligned}
\bm{E}'^m &= LN(\bm{E}^m + GRU(\bm{E}^m)), \\
\bm{H}^m &= LN(\bm{E}^m + \bm{E}'^m + FFN(\bm{E}'^m),
\end{aligned}
\end{equation}
where $\bm{H}^m \in \mathbb{R}^{n \times d_m},m \in \{a,v\}$, $LN$ denotes layer normalization, and $FFN$ denotes a feedforward neural network.

After encoding the local context for each modality, we obtain the sequence representations $\bm{H}^t$, $\bm{H}^a$, $\bm{H}^v$ for text, audio, and video.
We then apply three linear layers to map $\bm{H}^t$, $\bm{H}^a$, $\bm{H}^v$ to $\bm{H}'^t$, $\bm{H}'^a$, $\bm{H}'^v$ with the same dimension $d_h$.

\subsection{Graph Construction}
To enable M\textsuperscript{3}HG to fuse multi-scale semantic information across modalities, we construct a heterogeneous graph that represents both inter- and intra-utterance connections, as well as cross-modal interactions.
The structure of this heterogeneous graph can be denoted by $\mathcal{G}=(\mathcal{V},\mathcal{E},\mathcal{R})$, where $\mathcal{V}$ is the node set consisting of all graph nodes $v_i$, $\mathcal{R}$ is the relation set consisting of all relations $r_{ij}$ between any two nodes $v_i$ and $v_j$, and $\mathcal{E}$ is the edge set consisting of all edges represented as $(v_i,r_{ij},v_j)$.

\noindent\textbf{Nodes.} To explicitly model emotion and cause-related contexts in conversations, we model them as \textit{emotional context nodes} $N^e$ and \textit{causal context nodes} $N^c$, respectively.
To enable $\mathcal{G}$ to accurately perceive the conversation information, we model the whole conversation as a \textit{conversation node}.
Each utterance is represented by an \textit{utterance node}. 
Both the utterance node and conversation node are designed as Super-Nodes containing these modalities, denoted as $SN^u$ and $SN^d$, since they contain three modal features.
Therefore, $\mathcal{G}$ contains four types of nodes: $N^e$, $N^c$, $SN^u$ and $SN^d$.

$N^e$ and $N^c$ are first initialized with textual sequence representations $\bm{H}'^t$, then updated with contextual information from the other two modalities, which is described in Section~\ref{method-4.5}.
Each utterance Super-Node $SN^u =\{N^t,N^a,N^v\}$ is initialized using $\bm{H}'^t,\bm{H}'^a,\bm{H}'^v$.
The conversation node $SN^d=\{N_d^t,N_d^a,N_d^v\}$ is initialized by averaging $\bm{H}'^t,\bm{H}'^a,\bm{H}'^v$ to capture global information.

\noindent\textbf{Edges and Relations.} There are five types of Super-Edges connecting the aforementioned Super-Nodes: \textit{same speaker} ($r_{ss}$), \textit{different speaker} ($r_{ds}$), \textit{global connection} ($r_{gc}$), \textit{emotion connection} ($r_{ec}$) and \textit{cause connection} ($r_{cc}$).
The \textit{same speaker} edge connects the utterance Super-Nodes $SN^u$ from the same speaker.
Inspired by the work of \citet{DAG-ERC}, we define the local context as $K$ preceding utterances from the same speaker of $SN^u$, where $K$ is a hyper-parameter.
The \textit{different speaker} edge connects the utterance Super-Nodes within the local context from different speakers to $SN^u$.
The bidirectional \textit{global connection} edge connects all the utterance Super-Nodes $SN^u$s with the conversation Super-Node $SN^d$, facilitating the propagation of global contextual information.
The \textit{emotion connection} edge and the \textit{cause connection} edge 
connect $SN^u$ with its corresponding emotional context node $N^e$ and causal context node $N^c$, respectively.
They explicitly capture the emotion and cause context specific to each utterance.

M\textsuperscript{3}HG is the first MECTEC model capable of handling situations where \textbf{cause utterances appear after emotion utterances}, as each utterance is linked through the \textit{global connection} node.
The detailed experiments in Appendix~\ref{in-depth analysis} further validate this capability.
The pseudo-code of graph construction and a constructed graph for the conversation in Figure~\ref{fig:introduction figure} are provided in Appendix~\ref{appendix:pseudo-code} and Appendix~\ref{appendix-c.3}, respectively.
The graph construction process of M\textsuperscript{3}HG can be expressed as:
\begin{equation}
    \begin{aligned}
    \mathcal{G} &= (\mathcal{V}, \mathcal{E}, \mathcal{R}), \\
    \mathcal{V} &= \{ SN_i^u, N_i^e, N_i^c, SN^d \}_{1\leq i\leq n}, \\
    SN_i^u &= \{ N_i^t, N_i^a, N_i^v \},\\
    SN^d &= \{ N_d^t, N_d^a, N_d^v\}, \ \\
    \mathcal{R} &= \{ r_{ss}, r_{ds}, r_{gc}, r_{ec}, r_{cc} \}, \\
    \mathcal{E} &= \{(v_i, r_{ij}, v_j)\}, v_i, v_j \in \mathcal{V}, r \in \mathcal{R},
    \end{aligned}
\end{equation}
where superscripts $u,e,c,d$ denote node types,  and $m$ denotes three modalities.
Based on the constructed graph $\mathcal{G}$, the emotion and cause contexts are effectively modeled.

\subsection{Multi-scale Semantic Information Fusion}
\label{method-4.5}
Based on graph $\mathcal{G}$, we designed a comprehensive approach to integrate semantic information across different modalities and scales.
This mechanism is implemented in two levels: \textbf{intra-utterance fusion} which captures emotion and cause-related contexts within utterances, and \textbf{inter-utterance fusion} which propagates semantic information among utterances and conversation-level contexts.
Both levels leverage HGAT~\cite{HAN} to propagate and fuse semantic information through various meta-paths~\cite{HAN} within $\mathcal{G}$.
This ensures thorough updates to node features by integrating multi-scale semantic information.

The meta-paths in $\mathcal{G}$ are defined as:
\begin{equation}
\begin{aligned}
\Phi = \{\phi(v_i, r_{ij}, v_j)\}, v_i, v_j \in \mathcal{V}, \\
\phi(v_i, r_{ij}, v_j) = v_i \xleftrightarrow{r_{ij}} v_j, r_{ij} \in \mathcal{R},
\end{aligned}
\end{equation}
where $\phi(v_i, r_{ij}, v_j)$ represents all paths that connect node $v_i$ to node $v_j$ via edge type $r_{ij}$.

\noindent\textbf{Intra-utterance-level Fusion.} As shown in Figure~\ref{fig:overallframework}, for each utterance Super-Node $SN^u$, we perform intra-utterance-level fusion by integrating semantic information within the utterance.
We define the meta-path $\Phi_{intra}$ for  intra-utterance-level semantic fusion for $SN_n^u$ as:
\begin{equation}
\begin{aligned}
\Phi_{intra} = &\{\phi\left(N^{m_1}, N^{m_2}, r_{m_1,m_2}\right)\}\\
&\cup \{\phi\left(N^m, N^e, r_{m,e}\right)\} \\
&\cup \{\phi\left(N^m, N^c, r_{m,c}\right)\},
\end{aligned}
\end{equation}
where $m_1, m_2, m \in \{t, a, v\}$, $N^m$ represents the nodes of modality $m$ within the $SN^u$, and $r_{m_1,m_2}$ denotes the edges connecting $N^{m_1}$ and $N^{m_2}$. 
$r_{m,e}$ denotes edges connecting nodes $N^m$ to the emotional context nodes $N^e$, facilitating the aggregation of emotional contexts conveyed by different modalities within the utterance.
Similarly, $r_{m,c}$ represents the edges that connect $N^m$ to the causal context nodes $N^c$, enabling the aggregation of causal contexts.
$\Phi_{intra}$ effectively models the process of semantic information fusion in a single utterance.

Next, we incorporate node-level attention into $\Phi_{intra}$.
For each meta-path in $\Phi_{intra}$ and nodes $v_i\in \{N^m,N^e,N^c\}$, the importance of its neighbors $\mathcal{N}_i$ in $\Phi_{intra}$ is computed as:

\begin{small}
\begin{equation}
\alpha_{ij}^{\phi} = \frac{\exp\left(\sigma\left(\bm{a}_\phi^T \cdot \left[\bm{H}'_i \parallel \bm{H}'_j\right]\right)\right)}{\sum_{k \in \mathcal{N}_i^{\phi}} \exp\left(\sigma\left(\bm{a}_\phi^T \cdot \left[\bm{H}'_i \parallel \bm{H}'_k\right]\right)\right)}, \quad \phi \in \Phi_{intra},
\label{eq:11}
\end{equation}
\end{small}
where $\sigma$ denotes the activation function, and $\bm{a}_\phi$ is the node-level attention vector of meta-path $\phi$.
The node representation of $v_i$ based on meta-path $\phi$ is obtained by:
\begin{equation}
\bm{Z}_i = \sigma( \sum_{j \in \mathcal{N}_i^{\phi}} \alpha_{ij}^{\phi} \cdot \bm{H}'_j ).
\label{eq:12}
\end{equation}

This process yields the contextual features $\bm{Z}_i \in \mathbb{R}^{1 \times d_h}$ for nodes $v_i$ under the intra-utterance-level meta-paths $\Phi_{intra}$.

\noindent\textbf{Inter-utterance-level Fusion.} As illustrated in Figure~\ref{fig:overallframework}, for any two utterance Super-Nodes $SN_i^u$ and $SN_j^u$ in $\mathcal{G}$, along with the conversation Super-Node $SN^d$, we perform inter-utterance-level fusion by connecting $SN_i^u$ and $SN_j^u$ to $SN^d$, thereby integrating contextual information across utterances.
We define meta-paths $\Phi_{inter}$ for inter-utterance-level fusion between $SN^u$ and $SN^d$:
\begin{equation}
\begin{aligned}
\Phi_{inter} = &\{\phi (N_i^{m_1}, N_j^{m_2}, r_{m_1,m_2})\} \\
&\cup \{ \phi(N_i^m, N_d^m, r_{d,m})\} \\
&\cup \{ \phi(N_j^m, N_d^m, r_{d,m})\},
\end{aligned}
\end{equation}
where $m_1, m_2, m \in \{t, a, v\}$, $N_i^m$ and $N_j^m$ represent the nodes of modality $m$ inside $SN_i^u$ and $SN_j^u$, respectively, $r_{m_1,m_2}$ denotes the edges connecting $N_i^{m_1}$ and $N_j^{m_2}$, and $r_{d,m}$ represents the edges connecting $SN^u$s to $SN^d$ in modality $m$.

The utterance information from each modality can be passed to $SN^d$ though $\Phi_{inter}$, which accomplishes inter-utterance-level fusion between utterances.
As a result, $SN^d$ comprehensively integrates information across all three modalities.
The meta-path set $\Phi_{inter}$ models multimodal connections between utterances, enabling conversation information aggregated in $\mathcal{G}$.

Similar to Eq.~\ref{eq:11} and Eq.~\ref{eq:12}, the contextual representations of $SN^u$ and $SN^d$ are obtained under the meta-path $\Phi_{inter}$ by the node-level attention block.

After performing multi-scale semantic fusion with $\Phi_{intra}$ and $\Phi_{inter}$, we apply the semantic attention mechanism~\cite{HAN} to each node embedding $\bm{Z}_i$, integrating multi-scale semantic information from all three modalities.
Following \cite{MPEG}, each fusion iteration is followed by a position-wise feed-forward network (PFFN) layer, which updates node features through a non-linear transformation.
The emotional context node  representation $\bm{Z}_i^e$ and the causal context node feature representation $\bm{Z}_i^c$ can be obtained at the end of iterations of the multi-scale semantic fusion and PFFN layers.

\subsection{Emotion-cause classification}
For each utterance $\bm{U}_i$, its $\bm{Z}_i^e$ and $\bm{Z}_i^c$ are fed into the emotion-specific Multi-Layer Perceptron (Emotion MLP) and the cause-specific Multi-Layer Perceptron (Cause MLP) to predict its emotion category $\hat{y}_i^e$ and the cause indicator $\hat{y}_i^c$ which indicates whether $\bm{U}_i$ can be a cause utterance.
For each utterance pair $\bm{U}_i$ and $\bm{U}_j$, we compute a relative position encoding $RPE_{ij}$ to capture the positional relationship between $\bm{U}_i$ and $\bm{U}_j$.
We utilize the RBF kernel function~\cite{wei2020effective} to compute $RPE_{ij}$, which captures the relative positional relationships between utterances through a nonlinear relation.
$\bm{Z}_j^e$, $\bm{Z}_i^c$ and $RPE_{ij}$ are then concatenated and fed into a new MLP to determine whether $\bm{U}_i$ is the cause utterance of $\bm{U}_j$:
\begin{equation}
    {\hat{y}}_{ij}^{ec}=\sigma(MLP{(\bm{Z}_j}^e||{\bm{Z}_i}^c||RPE_{ij}).
\end{equation}
${\hat{y}}_{ij}^{ec}$ represent the binary classification logits indicating whether $\bm{U}_i$ is the cause of $\bm{U}_j$.
Based on ${\hat{y}}_{ij}^{ec}$, we can determine whether $\bm{U}_j$, $\bm{U}_i$ and $\hat{y}_j^e$ can form a true utter-cause-emotion triplet.
\renewcommand{\arraystretch}{0.8}
\begin{table*}[htbp]
    \centering
    \scriptsize
    \begin{tabular}{llllccccccccc}
        \toprule
        \textbf{Dataset} & \multicolumn{2}{c}{\textbf{Method}} & \textbf{Modality} & \textbf{Anger} & \textbf{Disgust} & \textbf{Fear} & \textbf{Joy} & \textbf{Sadness} & \textbf{Surprise} & \textbf{6 Avg.} & \textbf{4 Avg.} \\
        \midrule
        \multirow{13}{*}{ECF} 
        & Pipline & MC-ECPE-2steps\textsuperscript{$\triangle$} & T, A, V & 24.39 & 0.00 & 0.71 & 38.84 & 21.60 & 40.24 & 29.32 & 31.92 \\
        \cmidrule{2-12}
        & \multirow{5}{*}{E2E} 
        & ECPE-2D\textsuperscript{$\triangle$} & T & 25.13 & 0.00 & 0.00 & 41.25 & 21.62 & 43.24 & 30.80 & 33.55 \\
        & & RankCP & T & 28.29 & 12.03 & 3.52 & 38.69 & 22.17 & 37.67 & 30.58 & 32.48 \\
        & & UECA-Prompt\textsuperscript{$\triangle$} & T &  27.37 & 12.85 & 7.91 & 37.96 & 22.51 & 39.53 & 30.75 & 32.49 \\
        & & SHARK\textsuperscript{*} & T & 28.65 & 10.42 & 5.33 & 40.41 & 25.35 & 40.45 & 32.24 & 34.33 \\
        & & HiLo\textsuperscript{*}& T, A, V &-&-&-&-&-&-& 33.04 & 35.81 \\
        \cmidrule{2-12}
        & \multirow{1}{*}{LLMs}
        & GPT-4o (5-shots) & T & 28.49 & 17.76 & 12.35 & 31.11 & 27.27 & 33.89 & 29.13 & 30.30\\
       \cmidrule{2-12}
       & \multirow{4}{*}{}
       & \multirow{4}{*}{M\textsuperscript{3}HG (ours)} 
       & T & 34.47 & 18.17 & 12.72 & 43.28 & \underline{32.22} & 45.82 &37.46 & 39.95 \\
       & & & T, A & \underline{35.53} & \underline{18.71}  & \underline{17.07}  & \underline{47.73} & 30.97 & 46.72 & \underline{39.10} & \underline{40.97} \\
       & & & T, V & 34.05 & 18.18 & \textbf{19.57} & 46.23 & 32.10 & \textbf{48.50} & 38.90 & 40.72 \\
       & & & T, A, V & \textbf{36.08} & \textbf{23.33} & 9.88 & \textbf{49.03} & \textbf{32.41}  & \underline{47.46} & \textbf{40.07} & \textbf{41.96} \\

        \cmidrule{1-12}
        \multirow{13}{*}{MECAD} 
        & Pipeline & MC-ECPE-2steps & T, A, V & 28.43 & 0.00 & 0.23 & 22.45 & 27.67 & 45.14 & 22.01 & 24.83 \\
        \cmidrule{2-12}
        & \multirow{5}{*}{E2E} 
        & ECPE-2D & T & 28.12 & 0.00 & 0.56 & 24.30 & 28.01 & 35.87 & 25.32 & 28.54 \\
        & & RankCP & T & 29.79 & 12.50 & 3.06  & 21.79 & 29.31 & 32.36 & 26.29 & 28.32 \\
        & & UECA-Prompt & T &  28.54 & 12.12 & 5.32 & 20.84 & 29.67 & 34.17 & 25.91 & 27.87 \\
        & & SHARK & T & 30.22 & 10.16 & 4.10 & 25.84 & 30.21 & 34.59 & 27.58 & 29.99 \\
        & & HiLo\textsuperscript{*} & T, A, V &-&-&-&-&-&-& - & - \\
        \cmidrule{2-12}
        & \multirow{1}{*}{LLMs}
        & GPT-4o (5-shots) & T & 36.65 & 20.08 & 8.45 & 24.52 & 17.89 & 39.77 & 27.16 & 28.42 \\
       \cmidrule{2-12}
       & \multirow{4}{*}{}
       & \multirow{4}{*}{M\textsuperscript{3}HG (ours)} 
       & T & 35.85 & 18.05 & 15.38 & 25.95 & 29.13 & 42.11 & 30.81 & 32.55\\
       & & & T, A & \underline{37.29} & \underline{21.03} & \underline{15.89} & \underline{27.15}  &  30.34 & 42.78 & \underline{32.16} & \underline{33.73} \\
       & & & T, V & 36.91 & 20.48 & \textbf{16.91} & 25.47  & \underline{30.96} & \underline{43.14} & 31.95 & 33.52\\
       & & & T, A, V & \textbf{38.34} & \textbf{21.89} & 8.79 & \textbf{28.10} &  \textbf{31.17} & \textbf{43.29} & \textbf{32.82} & \textbf{34.59} \\
        \bottomrule
    \end{tabular}
    \caption{Performance comparison of different methods on the MECTEC task.
    $\triangle$ denotes the results are from~\cite{wang2023generative}. $*$ denotes the results are from the original paper~\cite{wang2023generative, li2024multimodal}. The best results and the second best results are in bold and underlined, respectively. Since HiLo~\cite{li2024multimodal} is not publicly available, we only report the results of HiLo on the ECF dataset.}
    \label{table:exp}
    \vspace{-2mm}
\end{table*} 

\subsection{Training}
We use Focal loss~\cite{ross2017focal} to cope with category imbalance in emotion-cause classification.
Specifically, the loss of both emotion prediction and cause utterance prediction and the emotion-cause pair prediction, can be expressed as:
\begin{equation}
    \mathcal{L}^\beta = -\frac{1}{N^\beta} \sum_{i=1}^{N^\beta} \alpha^\beta (1 - \hat{y}_i^\beta)^\gamma \log(\hat{y}_i^\beta),\\
    \beta \in \{e, c, ec\}
\end{equation}
where $\beta$ represents the task type, $N^\beta$ denotes the corresponding sample number of $\beta$, $\alpha^\beta$ is the category balancing factor, and $\gamma$ denotes the Focal loss modulation parameter.
These three training losses are optimized jointly during the training process.
\section{EXPERIMENTS}
\subsection{Experimental Settings}
We conduct extensive experiments on two MECTEC benchmark datasets, i.e., \textbf{ECF}~\cite{ecf} and \textbf{MECAD}, which both contain data of three modalities: text, audio, and video.
Similar to~\cite{ecf}, we evaluate the model's overall performance using the F1 score.
The F1 score is computed for utter-cause-emotion triplets within each emotion category separately.
Then the weighted average F1 score is calculated across all six emotion categories which is referred to \textit{6 Avg}.
In addition, as in~\cite{wang2023generative}, considering the data imbalance among different emotion categories, we also report the weighted average F1 scores for the four main emotion categories except \textit{Disgust} and \textit{Fear}, which is referred to \textit{4 Avg}. 
The implementation details of the experiment are given in Appendix~\ref{appendix-d}.

\subsection{Baselines}
Due to the limited research on the MECTEC task, representative approaches in related fields of Emotion Cause Pair Extraction (ECPE) and Emotion Cause Pair Extraction in Conversations (ECPEC) are considered.
The ECPE and ECPEC tasks aim to extract emotion-cause pairs from plain texts and conversations, respectively.

We compare our model with seven baselines:
1) \textbf{MC-ECPE-2steps}~\cite{ecf} is a two-step MECTEC architecture, which first extracts emotion utterances and cause utterances separately, and then performs pairing and filtering to identify emotion-cause pairs.
2) \textbf{HiLo}~\cite{li2024multimodal} is one of the SOTA approaches for the MECTEC task, which fully utilizes conversion information through a labeling constraint mechanism.
3) \textbf{ECPE-2D}~\cite{ECPE-2D} is an E2E framework for ECPE that uses 2D-Transformer to model the interactions of emotion-cause pairs.
4) \textbf{RankCP}~\cite{wei2020effective} is a GAT-based approach for ECPE to extract emotion-cause pairs by ranking.
5) \textbf{UECA-Prompt}~\cite{UECA-Prompt} is one of the SOTA methods for ECPE, which decomposes the task into multiple objectives and converts them into sub-prompts.
6) \textbf{SHARK}~\cite{wang2023generative} is the SOTA method for ECPEC that incorporates commonsense into GATs to improve the model's semantic understanding of emotions and causes.
7) \textbf{GPT-4o} is one of the most powerful large language models (LLMs) for open-domain conversations.
Details of prompts are provided in Appendix~\ref{gpt}.

\subsection{Experimental Results}
Table~\ref{table:exp} shows the experimental results of M\textsuperscript{3}HG and seven baseline models evaluated on the ECF dataset and the MECAD dataset.
Our model demonstrates an excellent performance both on the ECF dataset and the MECAD dataset.

\noindent\textbf{Results on the ECF dataset.} First of all, as shown in Table~\ref{table:exp}, among all the baseline models, the E2E approaches such as SHARK and HiLo deliver the best performance, indicating that the E2E framework is more effective compared to the two-step pipeline frameworks.
In contrast, M\textsuperscript{3}HG adopting three modalities outperforms the SOTA E2E model HiLo, with 21.28\% and 17.17\% improvement in \textit{6 Avg} and \textit{4 Avg} scores, respectively.
We attribute this improvement to M\textsuperscript{3}HG’s ability to effectively extract semantic information at inter-utterance and intra-utterance levels, which enables the model to accurately pair emotion utterances and cause utterances.
Specifically, in two challenging emotion categories which have limited training samples, i.e. \textit{Disgust} and \textit{Fear}, M\textsuperscript{3}HG also exhibits high performances.
For example, compared to GPT-4o, which achieved the second highest F1 scores in the \textit{Disgust} and \textit{Fear} categories, M\textsuperscript{3}HG shows improvements of 31.36\% and 58.46\%, respectively.

When only incorporating the text modality, the \textit{6 Avg} and the \textit{4 Avg} scores of M\textsuperscript{3}HG are 37.46 and 39.95.
When incorporating audio and video with the text modality separately, the performance of M\textsuperscript{3}HG is improved to 39.10, 38.90 of \textit{6 Avg} scores and 40.97, 40.72 of \textit{4 Avg} scores.
When incorporating all three modalities, M\textsuperscript{3}HG achieves the highest performance with 40.07 of \textit{6 Avg} scores and 41.96 of \textit{4 Avg} scores.
Meanwhile, it can be observed that M\textsuperscript{3}HG outperforms all the baseline models even when only using the text modality, demonstrating its superiority on the ECF dataset.

\noindent\textbf{Results on the MECAD dataset.} As shown in Table~\ref{table:exp}, M\textsuperscript{3}HG also achieves the highest results on the MECAD dataset.
Compared to the second best model SHARK, M\textsuperscript{3}HG adopting three modalities achieves the improvement of 19\% on the \textit{6 Avg} scores and 15.34\% on the \textit{4 Avg} score.
Furthermore, despite GPT-4o’s superior semantic comprehension abilities, its performance on the MECAD dataset remains suboptimal, with its \textit{6 Avg} score and \textit{4 Avg} score of 27.16 and 28.42.
Therefore, the few-shot-based LLM approach still struggles to effectively handle the MECTEC task.
As shown in Table~\ref{table:exp}, M\textsuperscript{3}HG exhibits a universal highest performance on the MECAD dataset, demonstrating the superiority and robustness of M\textsuperscript{3}HG when dealing with multi-conversation scenarios.

More detailed experimental results and the ablation study on M\textsuperscript{3}HG are presented in Appendix~\ref{more experimental results}.

\section{Conclusion}
In this work, we propose the first multimodal and multi-scenario Chinese emotion-cause analysis dataset, MECAD, for MECTEC and related emotion cause analysis tasks.
Compared to ECF, the only existing dataset for multimodal emotion-cause analysis, MECAD offers more diverse conversation scenarios.
It helps to enhance the generalizability and applicability of MECTEC models in complex social media environments.
Moreover, MECAD is a valuable resource for cross-cultural emotion analysis and recognition.
Furthermore, we propose a generalized MECTEC framework named M\textsuperscript{3}HG, which deeply extracts emotional and causal contexts, while effectively integrating semantic information across multiple granular levels.
Extensive experiments on the ECF dataset and the MECAD dataset demonstrate the superiority of our method compared to the existing state-of-the-art methods.
\section*{Limitations}
There are also some potential limitations in this work.
First, the process of emotional and causal context extraction does not integrate external knowledge, which limits the model's accuracy for emotion prediction and cause prediction.
In the future, we plan to integrate external knowledge into our model and leverage the advanced semantic extraction capabilities of current LLM technology to facilitate deeper and more precise emotion cause analysis. 
Second, M\textsuperscript{3}HG cannot handle excessively long conversations, as its input length is constrained by the language model used.
Furthermore, M\textsuperscript{3}HG may suffer from error propagation in the multimodal fusion process when emotion labels have uneven information across modalities.
This imbalance can lead to inaccurate predictions, especially when modalities conflict. This challenge is common in current multimodal models for emotion-cause analysis and suggests an area for future improvement. 

\section*{Ethical Considerations}
We did not use real-world conversations in our data collection because such conversations may violate the privacy of the speaker.
The effect of recruiting actors to play the roles is the same as in the TV series, but the scenes are not as diverse as in the TV series.
Therefore, we use TV series as the data source.
To further protect privacy, all data annotations were anonymized and de-identified, ensuring that our data collection adheres to ethical standards.
\bibliography{custom}

\appendix

\section{Dataset Statistics and Analysis of MECAD \label{appendix-a}}

To ensure the annotation quality of MECAD, we calculated Cohen's kappa~\cite{cohens-kappa} scores for every co-annotated data between two annotators, as shown in Figure~\ref{fig:annotation result}. 
The Cohen’s kappa~\cite{cohens-kappa} scores across all annotators are consistently around 0.6, indicating a good level of annotation consistency.

After the labeling was completed, we computed Cohen's kappa scores separately for data that were not co-labeled between the two labelers, as shown in Figure~\ref{fig:annotation result}.
Table~\ref{tab:MECAD statistics} lists some statistics of the MECAD dataset.
The dataset contains a total of 989 conversations, 10,516 utterances, and 8,077 emotion cause pairs from 56 different TV series, which ensures the size and diversity of the dataset.
Similar to M\textsuperscript{3}ED~\cite{M3ED}, we used TV-independent data segmentation to ensure the ability to validate model robustness as a benchmark dataset.
The average number of utterances and the average length of an utterance of a conversation are similar in the training, validation, and test sets.
At the same time, we can find that the average relative positions of the emotion cause pairs are all around 1, indicating that most of the emotions in the conversation are caused by the previous utterance.

\begin{figure}
\centering
  \includegraphics[width=0.45\textwidth]{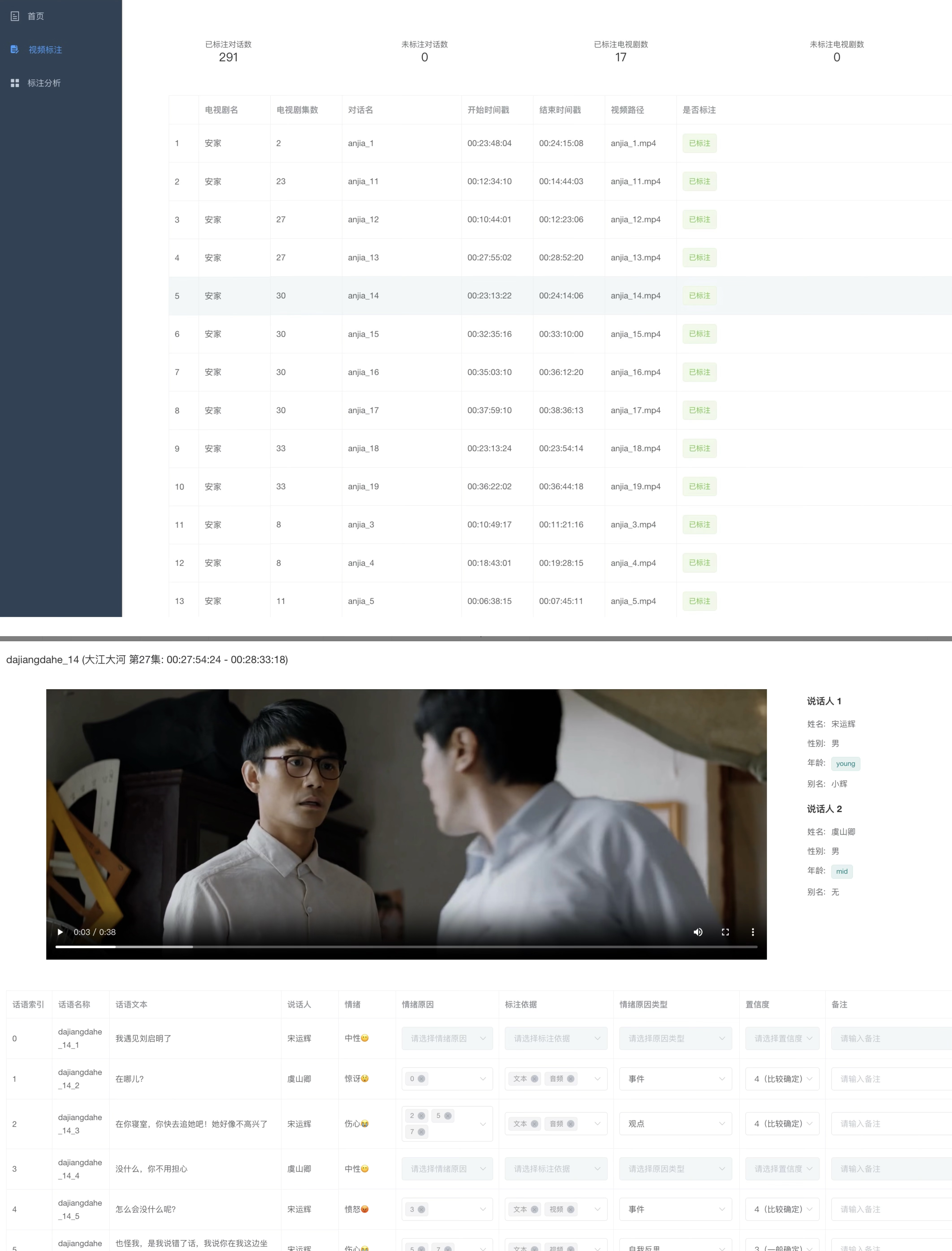}
  \caption{The interface of the developed online multimodal conversation emotion cause annotation toolkit.}
  \label{fig:annotation toolkit}
\end{figure}

\begin{figure}
\centering
  \includegraphics[width=0.45\textwidth]{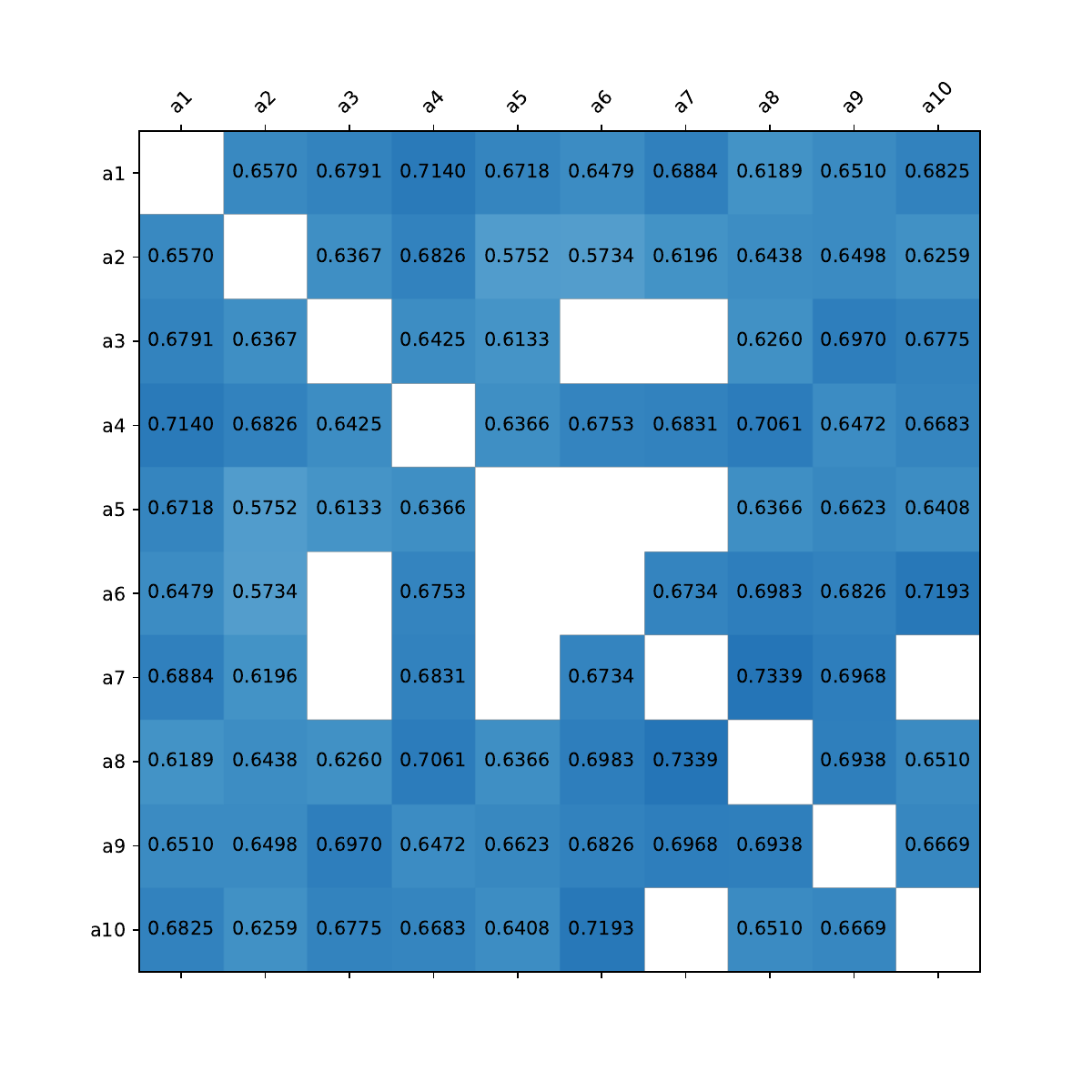}
  \caption{Schematic representation of Cohen's Kappa scores for the common labeled portion between every two annotators. A blank section indicates that there is no common annotation data between two annotators.}
  \label{fig:annotation result}
\end{figure}
\begin{table}
  \caption{MECAD statistics. \textit{Rel pos of ec pairs} denotes the relative position between emotion utterances and cause utterances in emotion-cause pairs.}
  \label{tab:MECAD statistics}
  \scriptsize
  \centering
  \begin{tabular}{lccc|c}
    \toprule
    Statistic & Train & Val & Test & Total\\
    \midrule
    \# TV series & 38 & 7 & 11 & 56\\
    \# conversations & 684 & 126  & 179 &  989 \\ 
    \# uttrs & 7,516 & 1,168  & 1,832 & 10,516\\
    \# spkrs & 421 & 87 & 118 & 626 \\
    Avg. uttrs/conversation & 10.99 & 9.27 & 10.24 & 10.63 \\
    Avg. uttr length & 18.30 & 18.80 & 18.15 & 18.33\\
    Avg. rel pos of ec pairs & 0.72 & 0.73 & 0.55 & 0.69\\
    Max. rel pos of ec pairs & 13 & 7 & 6 & 13\\
    Min. rel pos of ec pairs & -14& -5  & -9 & -14\\
    Emotion uttrs with cause & 4,526 & 743  & 1,062 & 6,331 \\
    ec pairs & 5,788  & 977  & 1,312 & 8,077 \\
  \bottomrule
  \end{tabular}
\end{table}
\label{appendix-b}
\begin{figure}
\centering
  \includegraphics[width=0.5\textwidth]{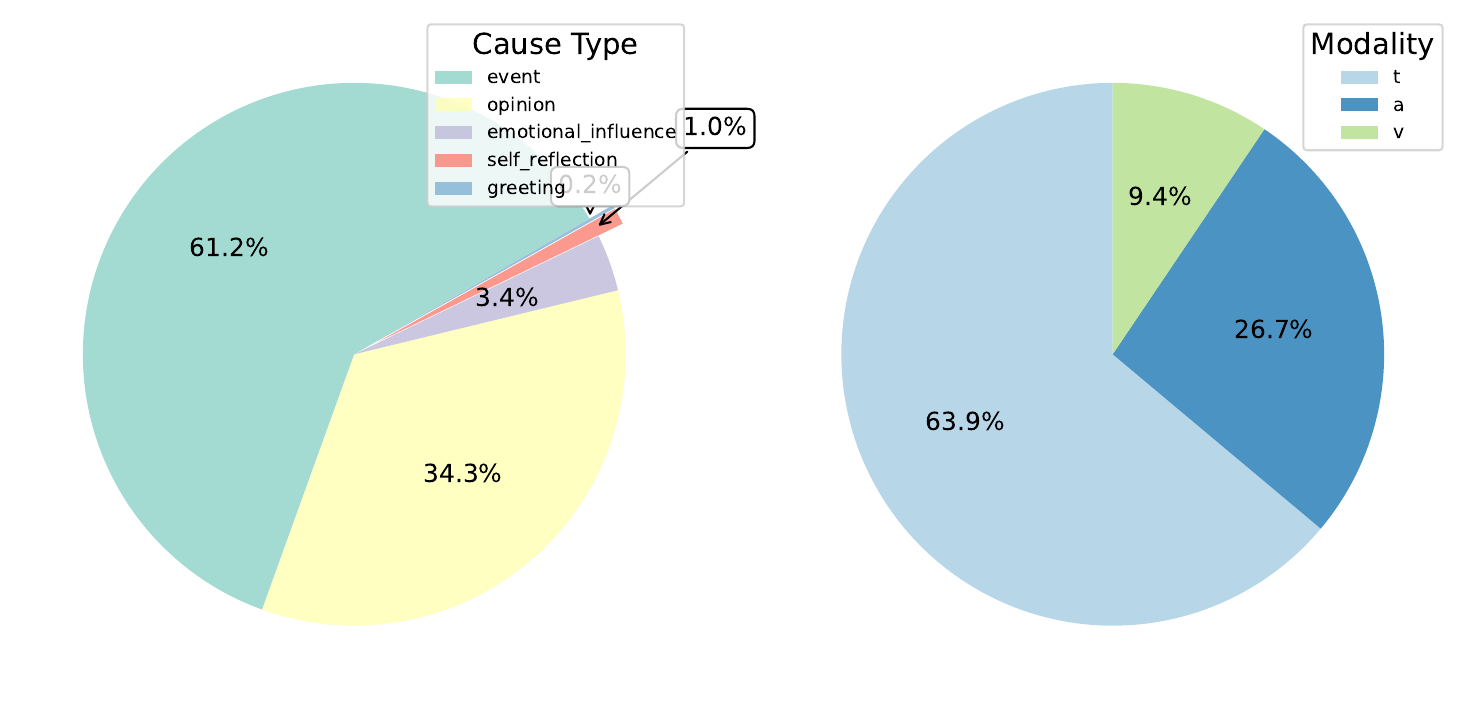}
  \caption{Percentage of five cause types in the MECAD dataset and percentage of modal basis for emotion cause inferences.}
  \label{fig:cause_type_distribution}
\end{figure}
We referred how the ECF~\cite{ecf} dataset categorizes the emotion causes and added a new category called \textit{Self Reflection}, which differs from the remaining four categories by indicating that emotions may be triggered by an individual's introspection or self-reflection, such as recollections of past events or worries about the future.
As shown in Figure~\ref{fig:cause_type_distribution}, the event type is the cause type with the largest share, indicating that most of the emotions are caused by specific events in the conversation.
Notably, 36.1\% of the causes of emotion in our dataset are reflected in both audio and video modalities, which exemplifies the need for multimodal scene studies.

\section{The Annotation Toolkit of MECAD}
\label{Appendix-toolkit}
To enhance annotation efficiency and accuracy, we developed an online multimodal conversation emotion cause annotation tool based on web technology\footnote{The annotation tool has been open-sourced at \url{https://github.com/redifinition/MECAD-MECTEC}}.
As illustrated in Figure~\ref{fig:annotation toolkit}, the toolkit’s homepage presents a list of conversations assigned to the corresponding annotators, along with the progress of their annotations.
The conversation annotation page displays speaker information, video segments, corresponding scripts, and configurable annotation items, enabling annotators to quickly and efficiently complete their annotations.

With flexible and modifiable web pages, researchers can utilize our annotation tools in dataset constructions for further multimodal sentiment analysis studies.
\begin{algorithm}
\caption{Super-Node-based Graph Constructuon for a Conversation}
\footnotesize
\begin{algorithmic}[1]
\State \textbf{Input}: the conversation $\{S_1:U_1, S_2:U_2, ..., S_N:U_N\}$, speaker identity $p(\cdot)$ satisfies $p(U_i) = S_i$, the direct context window $K$
\State \textbf{Output}: Super-Node-based M$^3$HG: $\mathcal{G} = (\mathcal{V}, \mathcal{E}, \mathcal{R})$
\State $\mathcal{V} \gets \{(SN_u^1, N_e^1, N_c^1), ..., (SN_u^N, N_e^N, N_c^N), SN_d^1\}$
\State $\mathcal{E} \gets \emptyset$
\State $\mathcal{R} \gets \{r_{ss}, r_{ds}, r_{gc}, r_{ec}, r_{cc}\}$
\For{$i \in \{2, 3, ..., N\}$}
    \State $c \gets 0$, $w \gets i - 1$
    \While{$w > 0$ and $c < K$} 
        \If{$p(U_w) = p(U_i)$}
            \State $\mathcal{E} \gets \mathcal{E} \cup \{(SN_u^w, SN_u^i, r_{ss})\}$ 
            \State $c \gets c + 1$
        \Else
            \State $\mathcal{E} \gets \mathcal{E} \cup \{(SN_u^w, SN_u^i, r_{ds})\}$ 
        \EndIf
        \State $w \gets w - 1$
    \EndWhile
\EndFor
\For{$i \in \{1, 2, ..., N\}$}
    \State $\mathcal{E} \gets \mathcal{E} \cup \{(SN_u^i, N_e^i, r_{ec})\}$ 
    \State $\mathcal{E} \gets \mathcal{E} \cup \{(SN_u^i, N_c^i, r_{cc})\}$ 
    \State $\mathcal{E} \gets \mathcal{E} \cup \{(SN_u^i, SN_d^i, r_{gc})\}$ 
\EndFor
\State \textbf{return} $\mathcal{G} = (\mathcal{V}, \mathcal{E}, \mathcal{R})$
\end{algorithmic}
\end{algorithm}
\section{\texorpdfstring{Design Details of M\textsuperscript{3}HG}{Design Details of M3HG}}
\subsection{Multimodal Feature Extracting}
\label{appendix-c.1}
\textbf{Text} : We splice all the textual modal utterances and the corresponding speakers in the conversation and add a number of special tokens to get the textual modal input sequence:
$X^t = \{<cls\_token> S_1:U_1^t, <sep\_token>, \dots, <cls\_token>S_n:U_n^t, <sep\_token> \}$,
where $<cls\_token>$ and $<sep\_token>$ denote the classification token and the separation token used in the pre-trained language model (PLM), respectively.
To allow conversations that exceed the maximum input sequence length of the PLM to retain as much contextual information as possible when they are fed into the PLM, we sequentially truncate the last tokens of the maximum-length utterances of the conversation during preprocessing until the maximum sequence length requirement of the PLM is met.
The input sequence $X^t$ is then fed into the PLM to obtain a sequential representation of the entire conversation:
\begin{equation}
\bm{I}^t = PLM(X^t),
\end{equation}

where $\bm{I}^t \in \mathbb{R}^{L \times d_t}$, $L$ is the length of the input sequence and $d_t$ is the hidden dimension of the PLM.
To obtain the sequence representation of each utterance, we make a weighted average of the sequence representations of the tokens of each conversation in $\bm{I}^t$ to obtain the sequence representation of each utterance $\bm{E}^t \in \mathbb{R}^{N \times d_t}$,where $N$ denotes the number of utterances of that conversation.
We selected Speaker-Aware RoBERTa (SA-RoBERTa)~\cite{SA-RoBERTa} as the PLM.

\textbf{Audio} : After resampling the audio to 16khz, we input it into an audio feature extraction model (AFE) to get a sequential representation of the audio modality of the conversation:
\begin{equation}
\bm{E}^a = AFE(X^a),
\end{equation}
where $\bm{E}^a \in \mathbb{R}^{n \times d_a}$, and $d_a$ is the hidden layer dimension of the audio feature extraction model. 
We choose Wav2Vec2.0~\cite{wav2vec2} as the audio feature extraction model.

\textbf{Video}: We first sample the video at equal intervals as a sequence of images over several frames to obtain the input sequence $\bm{X}^v$, $\bm{X}^v \in \mathbb{R}^{F \times d_f \times d_f }$ of the video modality, where $F$ is the number of sampled frames and $d_f$ is the size of the picture.
The image sequences are then fed into the video feature extraction model (VFE) to get a sequence representation of the video modalities:
\begin{equation}
\bm{E}^v = VFE(\bm{X}^v),
\end{equation}
where $\bm{E}^v \in \mathbb{R}^{n \times d_v}$ and $d_v$ is the hidden layer dimension of the video feature extraction model.
We select the pre-trained DenseNet\\~\cite{DenseNet} as the audio feature extraction model.

\subsection{Pseudo-code of Graph Construction}
\label{appendix:pseudo-code}

The pseudo-code of the graph construction process is shown in Algorithm~\ref{appendix:pseudo-code}.

\begin{table*}
  \centering
  \caption{Performance comparison of different methods for conversations with varying numbers of utterances. The best results and the second best results are in bold and underlined, respectively.}
  \label{tab:conversation length}
  \renewcommand{\arraystretch}{1.05}
  \small  
  {%
  \begin{tabular}{lccccccccc}
  \toprule
  \multirow{4}{*}{Method} & \multicolumn{4}{c}{\textbf{ECF}} & \multicolumn{4}{c}{\textbf{MECAD}} \\ \cmidrule(l){2-5} \cmidrule(l){6-9} & \multicolumn{2}{c}{num\_utt $\le$ 10} & \multicolumn{2}{c}{num\_utt $>$  10} & \multicolumn{2}{c}{num\_utt $\le$ 10} & \multicolumn{2}{c}{num\_utt $>$ 10} \\ \cmidrule(l){2-3} \cmidrule(l){4-5} \cmidrule(l){6-7} \cmidrule(l){8-9}
   & 6 Avg. & 4 Avg. & 6 Avg. & 4 Avg. & 6 Avg. & 4 Avg. & 6 Avg. & 4 Avg. \\ 
   \midrule
  RankCP  &  31.50 & 33.29 & 29.34 & 31.88 & 27.19 & 29.23 & 25.11 & 27.13 \\
  SHARK & 33.68 & 35.57 & 31.49 & 33.17 &  28.32 &   30.75 & 27.01 & 29.41 \\
  GPT-4o & 30.08 & 31.56 & 28.42 & 29.36 & 26.34 & 27.82 & 27.79 & 28.88 \\
  M\textsuperscript{3}HG (T) & \underline{39.18} & \underline{41.25} & \underline{36.18} & \underline{38.98} & \underline{31.95} & \underline{33.40} & \underline{29.94} & \underline{31.90} \\
M\textsuperscript{3}HG (T, A, V) & \textbf{41.95} & \textbf{40.42} & \textbf{38.67} & \textbf{41.09} & \textbf{33.76} & \textbf{35.21} & \textbf{32.10} & \textbf{34.12} \\
  \bottomrule
  \end{tabular}%
  }
\end{table*}
\subsection{An Example of the Graph construction}
\label{appendix-c.3}
\begin{figure}
  \includegraphics[width=0.9\linewidth]{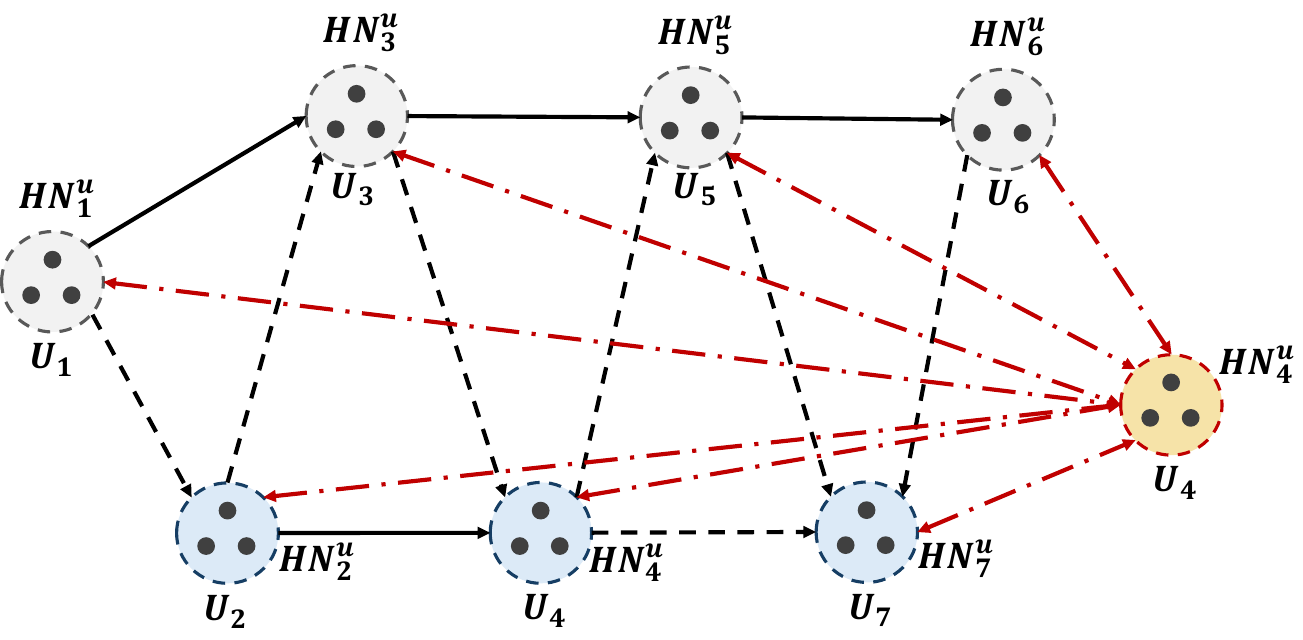}
  \caption{Super-Node-based edges and relations constructed from a conversation in MECAD with K = 1. The utterance Super-Nodes of the two speakers are shown in gray and blue, respectively. The black solid and dashed lines denote the Super-Edges between the same speaker and different speakers, respectively, and the red dotted lines denote the Super-Edges between the utterance Super-Nodes and the conversation Super-Nodes.}
  \label{fig:graph-example}
\end{figure}
If $K=1$, the graph constructed for the conversation in Figure~\ref{fig:introduction figure} is shown in Figure~\ref{fig:graph-example}.

\section{Implement Details of the Experiment}
For the ECF dataset, we use the pre-trained RoBERTa-large\footnote{\url{https://huggingface.co/FacebookAI/roberta-large}} model to initialize the feature extraction parameters of the text modality.
For audio modality, we use the wav2vec2-base-960h\footnote{\url{https://huggingface.co/facebook/wav2vec2-base-960h}} model and for video modality we use the DenseNet~\cite{DenseNet} model.
For the MECAD dataset, we use the chinese-roberta-wwm-ext-large\footnote{\url{https://huggingface.co/hfl/chinese-roberta-wwm-ext-large}} model for the initialization of textual modal features, the wav2vec2-large-chinese-zh-cn\footnote{\url{https://huggingface.co/wbbbbb/wav2vec2-large-chinese-zh-cn}} model for the extraction of audio modal features, and the DenseNet model is also applied to the video modal.
In our experiments, none of the parameters of the PLM were frozen.
During the construction of the graph, we set the hyperparameter $K$ to 3.
During training, we use the AdamW~\cite{AdamW} optimizer with batch size and learning rate set to 16 and 5e-6, respectively, and perform a parameter update after every two mini-batches.
Our model is trained for 50 epochs on the training set, and the checkpoints corresponding to the highest values of the weighted average F1 scores of the six emotions on the validation set are used as the results of the test set.
\label{appendix-d}

\begin{table*}
  \centering
  \caption{Performance comparison of different methods on four subtasks. The best results and the second best
results are in bold and underlined, respectively.}
  \label{tab:sub_task}
  \renewcommand{\arraystretch}{1.05}
  \scriptsize
  \begin{tabular}{lllcccccccccccc}
  \toprule
  \multirow{2}{*}{Dataset} & \multirow{2}{*}{Method} & \multicolumn{3}{c}{EP} & \multicolumn{3}{c}{ER} & \multicolumn{3}{c}{CE} & \multicolumn{3}{c}{EC} \\
  \cmidrule(lr){3-5} \cmidrule(lr){6-8} \cmidrule(lr){9-11} \cmidrule(lr){12-14}
   & & P & R & F1 & P & R & F1 & P & R & F1 & P & R & F1 \\
  \midrule
  
  \multirow{3}{*}{ECF} 
    & SHARK             & \underline{59.00} & 61.21 & \underline{60.74} & \underline{40.34} & \textbf{45.65} & \underline{42.83} & \underline{69.25} & 66.13 & 67.64 & \underline{50.12} & 46.31 & \underline{48.14} \\
    & GPT-4o(5-shots)   & 45.17 & \textbf{78.62} & 57.37 & 36.42 & 42.70 & 36.76 & 57.02 & \textbf{84.85} & \underline{68.21} & 32.90 & \textbf{61.13} & 42.78 \\
    & M\textsuperscript{3}HG (T)     & \textbf{71.36} & \underline{75.11} & \textbf{73.19} & \textbf{52.24} & \underline{45.63} & \textbf{46.60} & \textbf{72.32} & \underline{68.40} & \textbf{70.30} & \textbf{58.03} & \underline{52.05} & \textbf{54.88} \\
  
  \midrule
  
  \multirow{3}{*}{MECAD}
    & SHARK             & 69.30 & \underline{67.02} & \underline{68.14} & \underline{39.38} & \underline{36.93} & \underline{38.12} & 64.18 & 66.36 & 65.24 & \underline{49.02} & \underline{42.87} & \underline{45.74} \\
    & GPT-4o(5-shots)   & \underline{71.41} & 63.69 & 67.33 & 38.69 & 36.59 & 34.22 & \underline{65.03} & \underline{69.26} & \underline{67.08} & 39.68 & 41.77 & 40.70 \\
    & M\textsuperscript{3}HG (T)     & \textbf{72.34} & \textbf{67.84} & \textbf{70.02} & \textbf{43.35} & \textbf{40.98} & \textbf{41.66} & \textbf{66.12} & \textbf{70.24} & \textbf{68.12} & \textbf{54.82} & \textbf{46.42} & \textbf{50.27} \\
  
  \bottomrule
  \end{tabular}
\end{table*}

\section{\texorpdfstring{Supplementary Experimental Results of M\textsuperscript{3}HG}{Supplementary Experimental Results of M3HG}}
\label{more experimental results}
\subsection{Ablation Study}
\label{ablation study}
\textbf{Effect of different modules.} We conduct ablation studies to verify the effectiveness of different modules in M\textsuperscript{3}HG on the two datasets using \textit{6 Avg} and \textit{4 Avg} scores.
As shown in Table~\ref{tab:ablation study}, $w/o$ $N^e \& N^c$ indicates no use of emotional and causal context nodes in graph construction.
Consequently, emotion-cause pair prediction is performed directly based on the features of each utterance node.
$w/o$ \textit{inter-fusion} and $w/o$ \textit{intra-fusion} denote the absence of inter-utterance and intra-utterance multimodal fusion, respectively, during multi-scale semantic information fusion.
Our model outperforms the state-of-the-art baselines even without utilizing the previous three mechanisms.
Specifically, the performance of M\textsuperscript{3}HG degrades on both ECF and MECAD datasets when removing the emotional and causal context nodes, demonstrating the necessity of explicitly modeling the emotion and cause-related contexts.
Moreover, removing both intra-utterance and inter-utterance semantic fusion results in a drop in the model's performance, the former of which causes a more significant degradation.
It highlights the importance of effectively fusing semantic information at different scales within heterogeneous graphs, particularly within individual utterances.

\begin{table}
  \caption{Ablation results.}
  \scriptsize
  \label{tab:ablation study}
  \centering
  \begin{tabular}{clll}
    \toprule
    Dataset & Model & 6 Avg. &  4 Avg.\\
    \midrule
    \multirow{5}{*}{ECF} & M\textsuperscript{3}HG & \textbf{40.07} & \textbf{41.96} \\
    & $w/o$ all modules & $36.81_{(\downarrow3.26)}$ & $38.57_{(\downarrow3.39)}$ \\
    & \quad $w/o$ $N^e \& N^c$ & $38.13_{(\downarrow1.94)}$& $40.11_{(\downarrow1.85)}$\\
    & \quad $w/o$ \textit{inter-fusion} & $39.56_{(\downarrow0.51)}$ & $41.14_{(\downarrow0.82)}$ \\ 
    & \quad $w/o$ \textit{intra-fusion} &$39.12_{(\downarrow0.95)}$ & $40.86_{(\downarrow1.10)}$ \\ 
    \midrule
    \multirow{5}{*}{MECAD} & M\textsuperscript{3}HG & \textbf{32.82} & \textbf{34.59} \\
    & $w/o$ all modules & $30.37_{(\downarrow2.45)}$ & $32.27_{(\downarrow2.32)}$\\
    & \quad $w/o$ $N^e \& N^c$ & $30.94_{(\downarrow1.88)}$& $32.79_{(\downarrow1.80)}$\\
    & \quad  $w/o$ \textit{inter-fusion} & $32.57_{(\downarrow0.25)}$ & $33.91_{(\downarrow0.68)}$ \\ 
    & \quad $w/o$ \textit{intra-fusion} & $32.16_{(\downarrow0.66)}$ & $33.33_{(\downarrow1.26)}$ \\ 
  \bottomrule
  \end{tabular}
\end{table}

\begin{figure}
  \centering
  \includegraphics[width=0.7\linewidth]{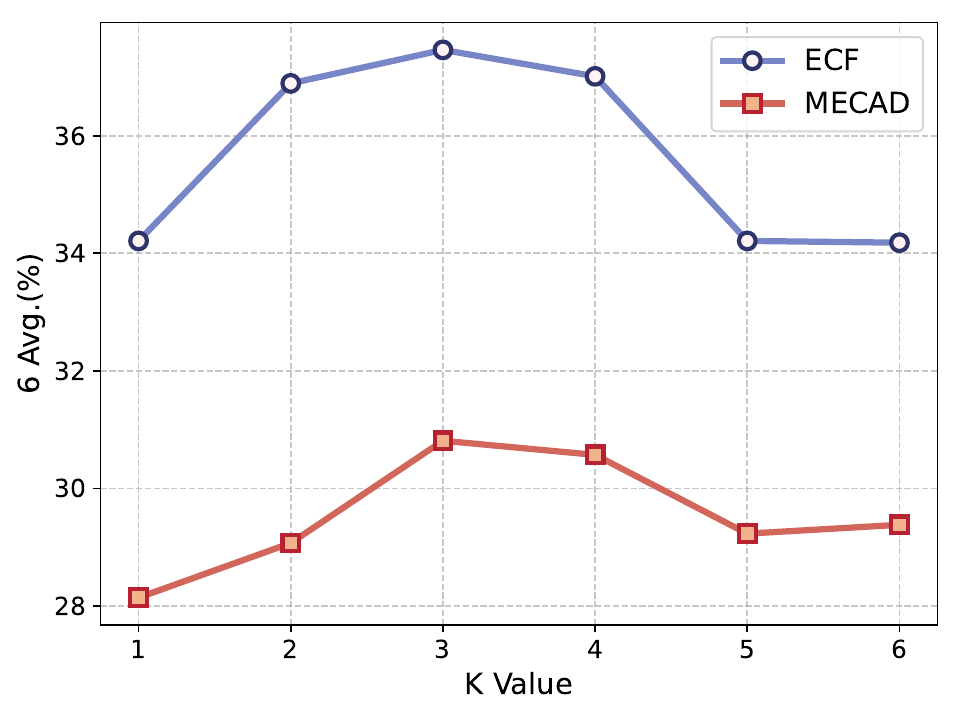}
  \caption{Results of M\textsuperscript{3}HG with various $K$ values.}
  \label{fig:k-value}
\end{figure}

\noindent \textbf{Effect of the hyperparameter $K$.}
The hyperparameter $K$ is closely related to the spatio-temporal complexity of the M\textsuperscript{3}HG's graph construction.
We vary the size of K (ranging from 1 to 6) to test its effect, and the result of the M\textsuperscript{3}HG on both datasets is shown in Figure~\ref{fig:k-value}. 
The performance of M\textsuperscript{3}HG on both datasets initially improves with increasing $K$ and then declines, with the best performance observed at $K = 3$.

\begin{figure*}
  \centering
  \includegraphics[width=\linewidth]{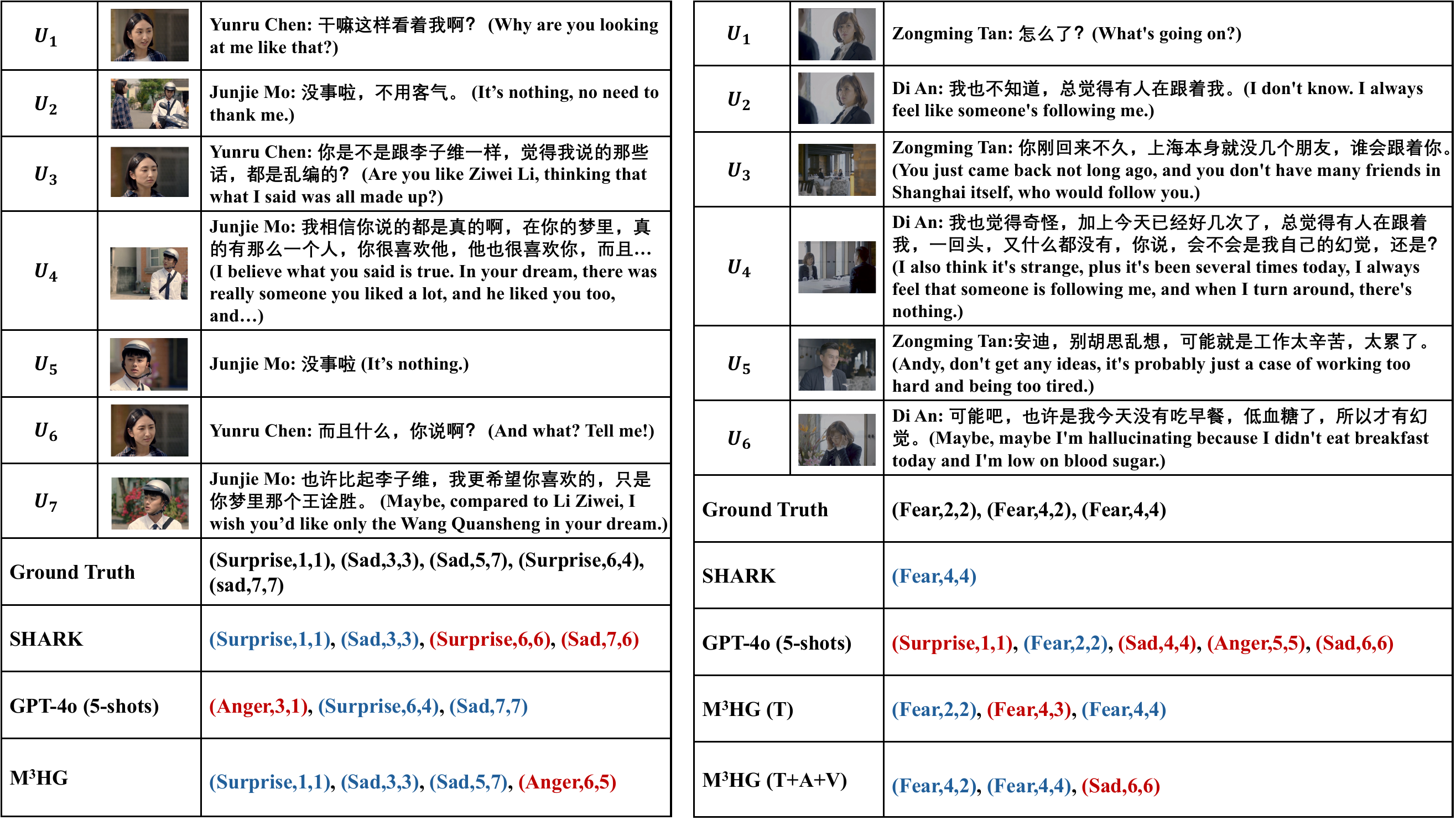}
  \caption{Comparison of utter-cause-emotion triplet on two test samples.}
  \label{fig:case study}
\end{figure*}

\subsection{In-Depth Analysis}
\label{in-depth analysis}
\noindent \textbf{The impact of conversation length.} 
To evaluate the performance of M\textsuperscript{3}HG in handling longer conversations, we present a comparison of the performance of M\textsuperscript{3}HG and other baseline models across conversations of varying lengths, as shown in Table~\ref{tab:conversation length}. 
We observe that M\textsuperscript{3}HG outperforms all baseline models in scenarios involving conversations with more than 10 utterances, which account for 42.65\% and 43.38\% of all conversations in the ECF and MECAD datasets, respectively.
In long conversations, baseline models, including GPT-4o, fail to effectively extract global contextual information, thereby missing a number of triplets.
Our model, through semantic fusion at different scales within multimodal heterogeneous graphs, effectively captures more triplets by extracting contextual information from long conversations.

\noindent \textbf{Model performance when cause utterance appears after emotion utterances.} 
A key challenge in the MECTEC task is when the cause of a speaker’s emotion is revealed later in the conversation, requiring the model to effectively capture and interpret the global context of the conversation.
To further emphasize the superior performance of M\textsuperscript{3}HG in handling cases where the cause utterance appears after the emotion utterance, we identified and filtered all such conversations from the ECF and MECAD datasets. The performance of M\textsuperscript{3}HG, compared with two other representative models, is shown in Table~\ref{tab:cause appears after emotion}.
M\textsuperscript{3}G demonstrates superior performance, while SHARK suffers a greater performance drop compared to M\textsuperscript{3}HG.
Although the performance drop for GPT-4o (5-shots) is less pronounced, its overall performance remains unsatisfactory.

\begin{table}
  \centering
  \renewcommand{\arraystretch}{1.05}
  \caption{Performance comparison of different methods for conversations in which cause utterance appears after emotion utterances. The best results and the second best results are in bold and underlined, respectively.}
  \label{tab:cause appears after emotion}
  \small  
  {%
  \begin{tabular}{lcccc}
  \toprule
  \multirow{3}{*}{Method} & \multicolumn{2}{c}{\textbf{ECF}} & \multicolumn{2}{c}{\textbf{MECAD}} \\ \cmidrule(l){2-3} \cmidrule(l){4-5} 
   & 6 Avg. & 4 Avg. & 6 Avg. & 4 Avg.  \\ 
   \midrule
  SHARK & 29.15 & 30.54 & 25.49 & 27.51 \\
  GPT-4o & 28.21 & 29.45 & 26.42 & 27.76 \\
  M\textsuperscript{3}HG (T) & \underline{35.48} & \underline{37.01} & \underline{29.18} & \underline{30.93} \\
M\textsuperscript{3}HG (T, A, V) & \textbf{38.25} & \textbf{40.01} & \textbf{31.27} & \textbf{33.09} \\
  \bottomrule
  \end{tabular}%
  }
\end{table}

\noindent \textbf{Model performance on four subtasks.}
To evaluate M\textsuperscript{3}HG’s performance more comprehensively, we define the following four subtasks:
\begin{itemize}[itemsep=0pt,parsep=0pt,topsep=0pt,partopsep=0pt]
  \item \textbf{Emotion Extraction (EP)}: Predict whether an utterance expresses an emotion (binary classification), same as SHARK.
  \item \textbf{Emotion Recognition (ER)}: Predict the emotion category of an utterance (multi-class classification).
  \item \textbf{Cause Extraction (CE)}: Predict whether an utterance is a cause utterance (binary classification), same as SHARK.
  \item \textbf{Emotion-Cause Pair Extraction (EC)}: Predict whether two utterances of a conversation form an emotion-cause pair (binary classification).
\end{itemize}

Table~\ref{tab:sub_task} demonstrates the performance comparison between M\textsuperscript{3}HG and other SOTA models across the four subtasks.
For the EP subtask, M\textsuperscript{3}HG performs the best across both datasets.
It is worth noting that GPT-4o (5-shots) achieves a high recall on the ECF dataset. 
This phenomenon can be attributed to the more pronounced label sparsity in the ECF dataset compared to MECAD. 
As a result, GPT-4o (5-shots) frequently predicts that an utterance carries emotion, leading to a higher recall.
For the ER subtask, M\textsuperscript{3}HG achieves the best results across both datasets.
This demonstrates M3HG’s ability to effectively extract the emotional context embedded in utterances.
For the CE subtask, M\textsuperscript{3}HG performs best, demonstrating the importance of integrating the cause prediction subtask into the model during training.
For the EC subtask, GPT-4o (5-shots) similarly exhibits high recall on the ECF dataset. 
This is due to the severe label sparsity problem in the ECF dataset, compared to MECAD, which leads GPT-4o to predict as many emotion-cause pairs as possible.

\begin{table*}[t]
    \renewcommand{\arraystretch}{1.05}
  \caption{An example of prompt for ChatGPT.}
  \scriptsize
  \label{tab:prompts}
  \centering
  \begin{tabular}{lll}
    \toprule
    \multirow{28}{*}{Input} & \multirow{8}{*}{Instruction} & You are an expert in sentiment analysis and identification of emotional causes. I will give you a conversation \\
    & & between multiple speakers. You are required to extract the utter-cause-emotion triplet for a given utterance.  \\
    & &  First, infer the emotion label for the utterance (select one from: Anger, Disgust, Fear, Joy, Sadness, Surprise \\
    & &   or Neutral). Then, identify the index(es) of the cause utterance(s) that triggered this emotion (the index  \\
    & & should represent the utterance(s) from the conversation that caused the emotion, and it must be non- \\
    & &  negative. Multiple indices  should be separated by commas). If the predicted emotion is Neutral, there is no \\
    & & corresponding cause utterance. The output should follow the format: emotion label, cause utterance indices. \\
    & &  Examples of the expected output format: Example 1: happy,3. Example 2: sad,3,4,5. Example 3: neutral. \\
    \cline{2-3}
    & \multirow{20}{*}{Demonstrations}& Input Conversation :\\
    & & \{ 1. Fang Sijin: First, change your clothes, then head to this address. A decoration company will be coming \\ 
    & &  over shortly. You’ll need to supervise their work and see how you can help. \}\\
    & & \{ 2. Zhu Shanshan: Wait, am I really responsible for this? I don’t know anything about decoration. \} \\
    & & \{ 3. Fang Sijin: You’ve been handing out flyers for two days now. Have you gotten any interested customers? \} \\ 
    & & \{ 4. Zhu Shanshan: But you only told me to distribute the flyers; you never ask for phone numbers! \} \\
    & & Candidate Utterances:  \\
    & & \{ 1. Fang Sijin: First, change your clothes, then head to this address. A decoration company will be coming \\ 
    & &  over shortly. You’ll need to supervise their work and see how you can help. \}\\
    & & \{ 2. Zhu Shanshan: Wait, am I really responsible for this? I don’t know anything about decoration. \} \\
    & & Target Utterance: \\
    & & \{ 2. Zhu Shanshan: Wait, am I really responsible for this? I don’t know anything about decoration. \} \\
    & & Target emotion labels and cause index(es): \\
    & & [Suprise, 1] \\
    & & Input Conversation : \\
    & & ...... \\
    & & Candidate Utterances: \\
    & & ...... \\
    & & Target Utterance: \\
    & & ...... \\
    & & Target emotion labels and cause index(es): \\
    & & ...... \\
    \cline{2-3}
     Output & output example & [Happy, 1, 2] \\
  \bottomrule
  \end{tabular}
\end{table*}

\subsection{Case Study}

To demonstrate the superiority and limitations of M\textsuperscript{3}HG, we present a case study that compares the prediction results of M\textsuperscript{3}HG with those of two other representative models (i.e. SHARK, GPT-4o (5-shots)), using two sample conversations from the MECAD dataset. 
As shown in Figure~\ref{fig:case study}, the first test sample demonstrates that M\textsuperscript{3}HG outperforms the other models in prediction accuracy, while GPT-4o exhibits the poorest performance.
This can be attributed to M\textsuperscript{3}HG’s use of a multimodal heterogeneous graph and a specially designed conversation super-node, which effectively captures global contextual information. These features enable M\textsuperscript{3}HG to more accurately handle scenarios where the cause utterance appears after the emotion utterance.

In the second sample, M\textsuperscript{3}HG (T+A+V) is less effective than M\textsuperscript{3}HG (T) in predicting Utterance 6 as “Sad” and Utterance 2 as “Neutral”. 
This is because the combination of text and context in Utterance 2 conveys the speaker’s worried and fearful mood, while the video and audio signals suggest a calmer demeanor.
This discrepancy likely caused M\textsuperscript{3}HG (T+A+V) to mispredict the emotions in this case. 
Nevertheless, M\textsuperscript{3}HG still outperforms all other baseline models, demonstrating its robustness and superior predictive capability even under challenging conditions.

\section{Prompt Design for ChatGPT}
\label{gpt}

We use the GPT-4o model of OpenAI public API (version up to May 13, 2024) and design a prompt elaborately to test the performance on the MECTEC task. The prompt (i.e., the input of ChatGPT) includes three parts:
\begin{itemize}
    \item \textbf{Instruction}. We use instructions to guide the ChatGPT on what it needs to do. Our instruction is as follows:

    \textit{You are an expert in sentiment analysis and identification of emotional causes. I will give you a conversation between two or more speakers. You need to extract the utter-cause-emotion triplet of the given utterance.}

    Meanwhile, we provide a detailed description of the output formats required for ChatGPT, as illustrated in Table~\ref{tab:prompts}.

    \item \textbf{Demonstrations} We achieve the few-shot in-context learning of ChatGPT by adding demonstrations. 
    We use the 5-shot in-context learning due to the limitations of the input length. 
    Each demonstration includes a conversation as input and a target utterance as the target for prediction.

\end{itemize}
Except for the aforementioned two parts, we also need to describe the conversations to be predicted and the corresponding target utterance.
An example is shown in Table~\ref{tab:prompts}.
\end{document}